\documentclass[journal]{IEEEtran}
\usepackage{amsmath,amssymb,amsfonts}
\usepackage{graphicx}
\usepackage{xcolor}
\usepackage{tikz}
\usetikzlibrary{
  arrows.meta,
  positioning,
  calc,
  fit,
  backgrounds,
  decorations.pathreplacing,
  shapes.geometric,
  patterns
}
\usepackage[ruled,vlined,linesnumbered]{algorithm2e}
\usepackage{booktabs}
\usepackage{multirow}
\usepackage{bm}
\usepackage{cleveref}

\definecolor{encblue}{RGB}{66,133,244}
\definecolor{decgreen}{RGB}{52,168,83}
\definecolor{teachorange}{RGB}{234,134,45}
\definecolor{lossred}{RGB}{204,51,63}
\definecolor{lightgray}{RGB}{240,240,240}
\definecolor{mirrpurple}{RGB}{128,90,213}
\definecolor{multiyellow}{RGB}{220,180,40}
\definecolor{darktext}{RGB}{40,40,40}

\newcommand{\Ltotal}{\mathcal{L}_{\text{total}}}

\newcommand{\Lssim}{\mathcal{L}_{\text{ssim}}}
\newcommand{\sg}{\operatorname{sg}}

\ifCLASSINFOpdf
\else
\fi

\usepackage{graphicx}
\usepackage{multirow}
\usepackage{booktabs}
\usepackage{amsmath}

\begin{document}
%
\title{IAML: Illumination-Aware Mirror Loss for Progressive Learning in Low-Light Image Enhancement Auto-encoders}
%
%
%

\author{Farida~Mohsen,
        Tala Zaim, Ali Al-Zawqari,
        Ali Safa,~\IEEEmembership{Member,~IEEE,}    and~Samir~Belhaouari,~\IEEEmembership{Senior Member,~IEEE}
\thanks{F. Mohsen, T. Zaim, A. Safa and S. Belhaouari are with the College of Science and Engineering, Hamad Bin Khalifa University, Doha, Qatar. emails: \{fmohsen, taza89388, asafa, sbelhaouari\}@hbku.edu.qa; }
\thanks{A. Al-Zawqari is with the ELEC department, Vrije Universiteit Brussel, Brussels, Belgium. emails: ali.mohammed.mohammed.al-zawqari@vub.be; }
\thanks{F. Mohsen developed the methods and performed the experiments. T. Zaim and A. Al-Zawqari contributed to the analysis of the results. A. Safa and S. Belhaouari supervised the project as Principal Investigators. All authors contributed to the writing of the manuscript.}
}

%
%

\markboth{Journal of \LaTeX\ Class Files,~Vol.~14, No.~8, August~2015}%
{Shell \MakeLowercase{\textit{et al.}}: Bare Demo of IEEEtran.cls for IEEE Journals}
%



\maketitle

\begin{abstract}
This letter presents a novel training approach and loss function for learning low-light image enhancement auto-encoders. Our approach revolves around the use of a teacher-student auto-encoder setup coupled to a progressive learning approach where multi-scale information from clean image decoder feature maps is distilled into each layer of the student decoder in a mirrored fashion using a newly-proposed loss function termed Illumination-Aware Mirror Loss (IAML). IAML helps aligning the feature maps within the student decoder network with clean feature maps originating from the teacher side while taking into account the effect of lighting variations within the input images. Extensive benchmarking of our proposed approach on three popular low-light image enhancement datasets demonstrate that our model achieves state-of-the-art performance in terms of average SSIM, PSNR and LPIPS reconstruction accuracy metrics. Finally, ablation studies are performed to clearly demonstrate the effect of IAML on the image reconstruction accuracy.    
\end{abstract}


%
\IEEEpeerreviewmaketitle
\section{Introduction}

\IEEEPARstart{I}{n} the past decade, low-light image enhancement using modern deep learning (DL) techniques has been an active area of research, with applications to nighttime vision for autonomous car navigation, remote sensing, photography enhancement and so on \cite{Zhou2023SurroundNet,Cai2023Retinexformer,Wu2025URetinexNetPP,10971192,10909219}. The use of DL models over traditional luminosity and exposure amplification has constituted a major leap forward within the field, enabling much higher fidelity image recovery even within extreme low-light conditions where conventional luminosity and exposure manipulation fails \cite{9609683,11020740,8955834,11202696}. 

Within this context, the use of convolutional \textit{auto-encoder} architectures has been widely adopted as one of the models of choice for low-light image enhancement \cite{9609683}. Certainly, auto-encoders constitute a natural choice for this application following their ability at reconstructing output images of the same type and dimension as their input image \cite{10149738}. Furthermore, the use of \textit{skip connections} \cite{7780459}, where the original input image is added as an additional input to subsequent layers, has also been proven to be critical for achieving high-accuracy image reconstruction. Popular models such as the U-Net architecture have therefore been utilized in many works related to image restoration, as a well-suited auto-encoder backbone equipped with skip connections and moderate compute complexity \cite{10.1007/978-3-319-24574-4_28}.

In addition to DL architectural choices, the choice of the \textit{loss function} used during training has been proven to be of crucial importance \cite{7797130}. Indeed, prior works have explored how the design of the loss function impacts the performance of low-light image enhancement models \cite{7797130}, in terms of standard benchmark metrics such as \textit{Structural Similarity Index Measure} (SSIM), \textit{Peak Signal-to-Noise Ratio} (PSNR) and \textit{Learned Perceptual Image Patch Similarity} (LPIPS) \cite{5596999,8578166}.

Following these realizations within the field, this paper focuses on the design of a novel loss function termed \textit{Illumination-Aware Mirror Loss} (IAML) for learning low-light enhancement auto-encoder models. Inspired by the growing use of teacher-student setups \cite{app14188109}, our proposed IAML loss is used alongside a custom teacher-student auto-encoder setup that distills information from clean image representations into each layer of the student's decoder sub-network, better guiding model learning during the training procedure. The contributions of this paper are the following: 

\begin{enumerate}
    \item We introduce a teacher-student paradigm for low-light image enhancement where the teacher processes clean images while the student enhances degraded ones,
    enabling implicit learning of clean image priors.

    \item We propose a novel IAML loss that aligns student decoder feature maps with clean teacher decoder ones across all layers, facilitating multi-scale feature learning and distillation. 

    \item We perform extensive benchmarking and ablation studies to clearly demonstrating the usefulness of IAML and we show that our proposed approach outperforms more than a dozen prior techniques on three popular benchmark datasets. 
    
\end{enumerate}

This paper is organized as follows. The proposed methods are introduced in Section \ref{proposedmeth}. Experimental studies are presented in Section \ref{experimentsst}. Finally, conclusions are provided in Section \ref{concss}. 

\begin{figure*}[h]
\centering
\includegraphics[scale=0.4]{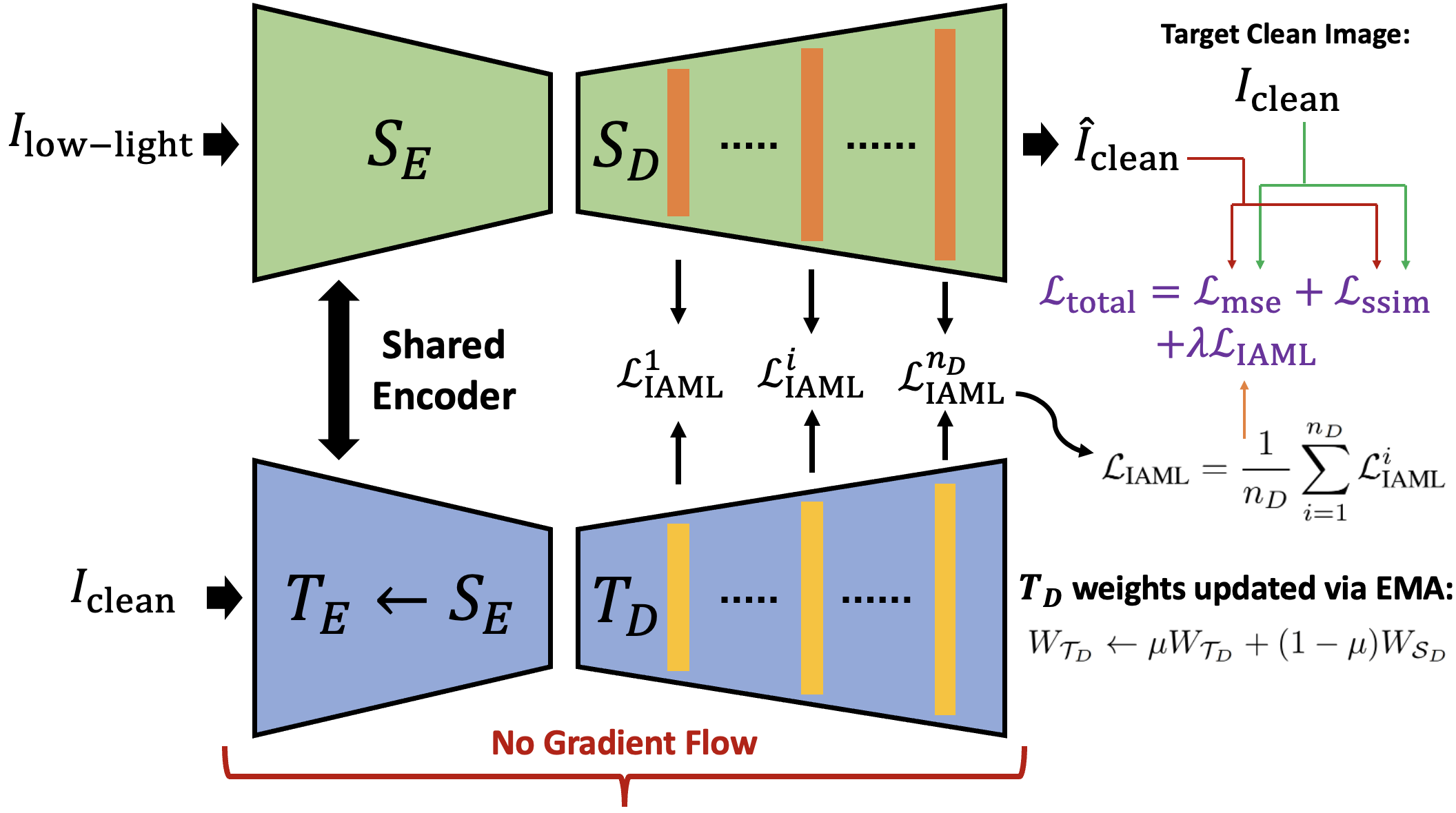}
\caption{Proposed teacher-student auto-encoder setup for low-light image enhancement. The student network $\mathcal{S}$ receives as input the degraded low-light images $I_{\text{low-light}}$ and seeks to recover the clean images $\hat{I}_{\text{clean}}$. The teacher network $\mathcal{T}$ shares the same encoder as $\mathcal{S}$ (updated as $\mathcal{T}_E\xleftarrow[]{} \mathcal{S}_E$ during the training iterations) and updates its decoder $\mathcal{T}_D$ via an exponential moving average using the weights of the student decoder $\mathcal{S}_D$ (no gradient-based learning in $\mathcal{T}$). The teacher takes as input the clean target images $I_{\text{clean}}$, leading to clean multi-scale feature maps at each layer of $\mathcal{T}_D$. These clean feature maps are then used to guide the learning of the layers within $\mathcal{S}_D$ in a mirrored fashion using the proposed Illumination-Aware Mirror Loss (IAML).}
\label{fig11}
\end{figure*}

\section{Proposed Method}
\label{proposedmeth}
This section describes our proposed method for low-light image enhancement, based on the formulation of a multi-level mirror learning scheme between a \textit{teacher} encoder network and a \textit{student} auto-encoder which learns to output the enhanced image from the degraded low-light input image.

\subsection{Teacher-student auto-encoder setup}
\label{setupnets}
Our proposed neural network setup is composed of both a teacher and a student auto-encoder, as shown in Fig. \ref{fig11}. First, we set up a \textit{student} auto-encoder $\mathcal{S}$ which always takes as input the degraded low-light images $I_{\text{low-light}}$ and aims to reconstruct the corresponding target clean images $I_{\text{clean}}$.
\begin{equation}
    \min ||\hat{I}_{\text{clean}} -  I_{\text{clean}}||_2 \hspace{5pt} \text{given} \hspace{5pt} \hat{I}_{\text{clean}} = \mathcal{S}(I_{\text{low-light}}) 
\end{equation}

We respectively denote by $\mathcal{S}_E$ and $\mathcal{S}_D$ the encoder and decoder parts of $\mathcal{S}$. Then, in addition to the student auto-encoder, we set up a \textit{teacher} auto-encoder $\mathcal{T}$ which strictly takes as input the target clean images $I_{\text{clean}}$ and seeks to re-produce them as its output. Crucially, the encoder part of $\mathcal{T}$ noted $\mathcal{T}_E$ is an exact copy of the student encoder $\mathcal{S}_E$ (shared encoder), with the difference that the backpropagation of gradients is \textit{disabled} so that the clean images given as input to the teacher encoder $\mathcal{T}_E$ never impact the learning of this shared encoder block:
\begin{equation}
    \mathcal{T}_E \xleftarrow[]{} \mathcal{S}_E
\end{equation}

Now, regarding the teacher decoder $\mathcal{T}_D$, both the student and teacher encoders are initialized with the same random weights at the beginning of the training process:
\begin{equation}
    \mathcal{T}_D^0, \mathcal{S}_D^0 \xleftarrow[]{} \text{random init} 
\end{equation}
Then, during the learning process, gradients are allowed to flow within the \textit{student} decoder while being \textit{blocked} for the \textit{teacher} decoder once again. Instead, at each learning iteration, the teacher decoder $\mathcal{T}_D$ is updated following an \textit{exponential moving average} (EMA) scheme as follows \cite{10.1007/978-3-031-73010-8_16}:
\begin{equation}
    W_{\mathcal{T}_D} \xleftarrow[]{} \mu W_{\mathcal{T}_D} + (1-\mu) W_{\mathcal{S}_D}
    \label{exponenat}
\end{equation}
where $\mu$ is the EMA coefficient, $W_{\mathcal{T}_D}$ denotes the teacher decoder weights and $W_{\mathcal{S}_D}$ denotes the student decoder weights (where gradient-based update occurs). During training, the multi-scale feature representations within $\mathcal{S}_D$ will undergo an \textit{alignment} process with regard to the representations within $\mathcal{T}_D$, corresponding to feature maps associated with the clean target images $I_{\text{clean}}$ that the student need to produce. This is done through the formulation of a novel loss function strategy, which will be explained in Section \ref{mirrorloss}.

The combination of the shared encoder $\mathcal{S}_E = \mathcal{T}_E$ processing both the degraded low-light and clean inputs, together with the EMA-driven teacher decoder $\mathcal{T}_D$ creates a \emph{self-distillation loop}: the teacher receives as input images in the clean domain and gradually converges to produce \textit{ideal} intermediate representations, which the student processing within the degraded domain is trained to mirror. The EMA-based update of $\mathcal{T}_D$ ensures smooth, non-collapsing targets. Gradient stopping prevents the teacher from being directly optimized, preserving the asymmetric role of the two branches.

\subsection{Illumination-aware mirror loss}
\label{mirrorloss}

The goal of our proposed IAML is to force each feature map produced within $\mathcal{S}_D$ to align itself with the corresponding feature map of $\mathcal{T}_D$ which processes the target clean images. Doing so, layer-wise and \textit{multi-scale} information about the feature representations of clean images are injected at each scale level of $S_D$, better guiding it during learning towards the generation of clean images from its degraded low-light inputs.

Crucially, our proposed mirror loss is made \textit{illumination-aware}. Indeed, because the student and teacher model decode features from two different
illumination domains, their intermediate activations occupy different dynamic ranges. Hence, a naive $\ell_1$ or $\ell_2$ matching between corresponding student and teacher feature maps would be dominated by domain-level magnitude differences rather than structural discrepancies. 

To alleviate this, we adopt a \textit{weighted loss} approach taking into account the luminance of each pixel in the input images as follows. Given the low-light input $I_{\text{low-light}}$, we compute a luminance map $L_\mathbf{p}$ per pixel $\mathbf{p}$ via the standard RGB-to-luminance coefficients \cite{8100110}:
\begin{equation}\label{eq:lum}
  L_\mathbf{p}
    = 0.299\, I_{\text{low-light},\mathbf{p}}^R
    + 0.587\, I_{\text{low-light},\mathbf{p}}^G
    + 0.114\, I_{\text{low-light},\mathbf{p}}^B
\end{equation}
Then, $L_\mathbf{p}$ is min-max normalized to $\tilde{L}_\mathbf{p} \!\in\! [0,1]$, and converted to an \textit{emphasis weight matrix}:
\begin{equation}\label{eq:weight}
  W_\mathbf{p}
    = 1 + \beta\,(1 - \tilde{L}_\mathbf{p}),
    \qquad \beta = 0.6,
\end{equation}
so that darker regions receive weight up to $1\!+\!\beta = 1.6$ while
well-lit regions receive the base weight of~$1.0$. Later on, upon feeding an input sample, its corresponding $W_\mathbf{p}$ can be resized to match the size of each feature scale $i$, noted as $W^i$. 

In addition, we first apply \textit{standard normalization} to the \textit{flattened} student and teacher feature maps $f_S^i,f_T^i$ for all decoder scale levels $i$ as follows:
\begin{equation}\label{eq:instnorm}
  \tilde{f}_{S,T}^i \xleftarrow[]{} \frac{\tilde{f}_{S,T}^i - \mu(\tilde{f}_{S,T}^i)}{\sigma(\tilde{f}_{S,T}^i + \epsilon)},
  \qquad \epsilon = 10^{-6},
\end{equation}
where $\mu$ and $\sigma$ respectively denote the mean and standard deviation operators (and $\epsilon$ avoids division by zero). Doing so, the proposed IAML at decoder layer level $i$ is then defined as:
\begin{equation}\label{eq:mirror}
  \mathcal{L}_{\text{IAML}}^{i}
    = \Bigl\langle W^i \odot
      \bigl|\,\tilde{f}_{S}^i - \sg(\tilde{f}_{T}^i)\,\bigr|
    \Bigr\rangle,
\end{equation}
where $\langle\cdot\rangle$ denotes spatial averaging, $\odot$ is the element-wise multiplication with broadcasting over the channel dimension and $\sg(\cdot)$ denotes the blocking of the
gradient flow associated with the teacher data. Finally, the total IAML is computed as the average across all decoder levels $i=1,...,n_D$:
\begin{equation}\label{eq:mirror_total}
  \mathcal{L}_{\text{IAML}} = \frac{1}{n_D}\sum_{i=1}^{n_D} \mathcal{L}_{\text{IAML}}^{i}.
\end{equation}

\subsection{Global loss function}

During training, the final loss is computed as a combination of a per-pixel reconstruction loss and the proposed IAML:
\begin{equation}\label{eq:total}
  \Ltotal
    = \underbrace{
        \mathcal{L}_{\text{mse}} + \Lssim
      }_{\text{per-pixel reconstruction}}
    + \underbrace{
        \lambda \mathcal{L}_{\text{IAML}}
      }_{\text{feature map distillation}}
\end{equation}
where $\mathcal{L}_{\text{mse}}$ is the standard mean square error between the student decoder output $\hat{I}_{\text{clean}}$ and the target clean image $I_{\text{clean}}$, $\Lssim = 1 - \text{SSIM}(\hat{I}_{\text{clean}}, I_{\text{clean}})$ is the
\textit{structural similarity} loss \cite{7797130} and $\lambda$ is a hyper-parameter defining the strength of the proposed IAML contribution.

\begin{table}[t]
\centering
\small
\setlength{\tabcolsep}{6pt}
\caption{Quantitative comparison of methods on the LOL-v1 dataset. Best scores are in bold.}
\begin{tabular}{lccc}
\hline
Method & SSIM$\uparrow$ & PSNR$\uparrow$ & LPIPS$\downarrow$ \\
\hline
LIME~\cite{Guo2016LIME} & 0.564 & 16.74  & 0.350 \\
SRIE~\cite{Fu2016WeightedVariational} & 15.12 & 0.569 & 0.340 \\
BVIF~\cite{Yang2020BVIF} & 0.561 & 16.61  & 0.286 \\
BIMEF~\cite{Ying2017BIMEF} & 0.566 & 14.80  & 0.326 \\
ZeroDCE++~\cite{Li2021ZeroDCEpp} & 0.595 & 16.72  & 0.335 \\
EnlightenGAN~\cite{Jiang2021EnlightenGAN} & 0.650 & 17.48  & 0.322 \\
RetinexNet~\cite{Wei2018RetinexNet} & 0.560 & 16.77 & 0.474 \\
KinD~\cite{Zhang2019KinD} & 0.790 & 20.86  & 0.207 \\
MIRNet~\cite{Zamir2020MIRNet} & 0.830 & \textbf{24.14}  & 0.131 \\
UFormer~\cite{Wang2022UFormer} & 0.771 & 16.36  & 0.321 \\
Restormer~\cite{Zamir2022Restormer} & 0.823 & 22.43 & 0.141 \\
LANet~\cite{Yang2023LANet} & 0.810 & 21.71 & 0.101 \\
SurroundNet~\cite{Zhou2023SurroundNet} & 0.853 & 23.84  & 0.113 \\
Retinexformer~\cite{Cai2023Retinexformer} & 0.816 & 23.69 & 0.140 \\

SCFR-WLMB~\cite{lv2025low} & 0.811 & 23.270  & -- \\
LightenDiffusion~\cite{jiang2024lightendiffusion} & 0.811 & 19.977  & 0.178 \\

\hline


\textbf{IAML (ours)} & \textbf{0.876} & \underline{23.84}  & \textbf{0.0848} \\
\hline
\end{tabular}
\label{lol1}
\end{table}

\begin{table*}[]
\centering
\small
\setlength{\tabcolsep}{5pt}
\caption{Quantitative comparison of methods on the LOL-v2-real and LOL-v2-synthetic datasets. Best scores are in bold.}
\begin{tabular*}{\textwidth}{@{\extracolsep{\fill}}lcccccc}
\hline
\multirow{2}{*}{Method} 
& \multicolumn{3}{c}{LOL-v2-real} 
& \multicolumn{3}{c}{LOL-v2-synthetic} \\
\cmidrule(lr){2-4} \cmidrule(lr){5-7}
 & SSIM$\uparrow$ & PSNR$\uparrow$ & LPIPS$\downarrow$
 & SSIM$\uparrow$ & PSNR$\uparrow$ & LPIPS$\downarrow$ \\
\hline

LIME~\cite{Guo2016LIME} & 0.469 & 15.19  & 0.415 & 0.776 & 16.85  & 0.675 \\
SRIE~\cite{Fu2016WeightedVariational} & 0.556 & 16.12 & 0.332 & 0.617 & 14.51  & 0.196 \\
BVIF~\cite{Yang2020BVIF} & 0.546 & 16.43 & 0.317 & 0.717 & 17.24  & 0.198 \\
BIMEF~\cite{Ying2017BIMEF} & 0.613 & 16.51  & 0.307 & 0.616 & 14.09  & 0.193 \\
ZeroDCE++~\cite{Li2021ZeroDCEpp} & 0.571 & 18.75  & 0.313 & 0.829 & 18.04  & 0.168 \\
EnlightenGAN~\cite{Jiang2021EnlightenGAN} & 0.617 & 18.23  & 0.309 & 0.734 & 16.57  & 0.212 \\
RetinexNet~\cite{Wei2018RetinexNet} & 0.567 & 15.47  & 0.365 & 0.798 & 17.13  & 0.754 \\
KinD~\cite{Zhang2019KinD} & 0.641 & 14.74  & 0.375 & 0.578 & 13.29  & 0.435 \\
MIRNet~\cite{Zamir2020MIRNet} & 0.820 & 20.02  & 0.317 & 0.876 & 21.94  & 0.112 \\
UFormer~\cite{Wang2022UFormer} & 0.771 & 18.82  & 0.347 & 0.871 & 19.66  & 0.107 \\
Restormer~\cite{Zamir2022Restormer} & 0.827 & 19.94  & 0.191& 0.830 & 21.41  & 0.066 \\
LANet~\cite{Yang2023LANet} & 0.780 & 19.21 & 0.178 & 0.882 & 22.54  & 0.069 \\
SurroundNet~\cite{Zhou2023SurroundNet}  & 0.807 & 20.18 & 0.144 & 0.900 & 23.88  & 0.068 \\
Retinexformer~\cite{Cai2023Retinexformer}  & 0.814 & 21.43 & 0.184 & 0.930 & 24.16  & 0.061 \\
URetinex-Net++~\cite{Wu2025URetinexNetPP} & 0.811 & 20.55  & 0.143 & 0.899 & 24.02 & 0.067 \\

SCFR-WLMB~\cite{lv2025low} 
& 0.834 & 21.430  & -- 
& 0.917 & \textbf{25.830} & -- \\

LightenDiffusion~\cite{jiang2024lightendiffusion}
& 0.853 & \textbf{22.831}  & 0.167   
& 0.867 & 21.523  & 0.157 \\ 

\hline
\textbf{IAML (ours)}
& \textbf{0.855} & \underline{21.45}  & \textbf{0.15} & \textbf{0.932} & \underline{24.82}  & \textbf{0.058}
\\

\hline
\end{tabular*}
\label{lol2}
\end{table*}

\section{Experimental setup}
\label{experimentsst}
\subsection{Model architecture and training procedure}
As auto-encoder backbone, we use the popular U-Net architecture with CBAM attention~\cite{woo2018cbam} used at each scale. As detailed in Section \ref{setupnets}, both the student and teacher networks share this same architecture. During training, images are randomly cropped to $256\!\times\!256$ patches with horizontal and vertical flipping augmentation. We use the Adam optimizer with an initial learning rate of $2\!\times\!10^{-4}$, $\beta_1\!=\!0.9$, $\beta_2\!=\!0.999$, and
a cosine-annealing schedule over 500 epochs with batch size~8. The EMA momentum in (\ref{exponenat}) is fixed at $\mu\!=\!0.999$ throughout training. In addition, grid search has been performed on the hyper-parameter $\lambda$ in (\ref{eq:total}) and the best results corresponding to $\lambda=0.8$ are reported. All experiments are carried using an NVIDIA RTX 4090 GPU.

\subsection{Benchmark Datasets}
We evaluate our proposed approach on three \textit{widely adopted} low-light benchmarks: \texttt{LOL-v1}, \texttt{LOL-v2-Real} and \texttt{LOL-v2-Synthetic}. First, \texttt{LOL-v1}~\cite{Wei2018RetinexNet} contains 485 training and 15 test
image pairs captured under real low-light conditions. Then,
\texttt{LOL-v2-Real}~\cite{yang2021sparse} extends the \texttt{LOL-v1} protocol with 689 training and 100 test pairs captured in real-world settings. Finally, \texttt{LOL-v2-Synthetic}~\cite{yang2021sparse} provides 900 training and 100 test pairs generated via synthetic degradation pipelines. All three datasets supply spatially aligned low-light / normal-light pairs enabling supervised training and reliable quantitative evaluation.

\subsection{Benchmark results}

During the benchmarking process, we use three widely used and standard similarity measurement scores: SSIM (the higher the better), PSNR (the higher the better) and LPIPS (the lower the better) \cite{5596999,8578166}.

On \texttt{LOL-v1}, our method achieves state-of-the-art perceptual quality, with the highest SSIM and lowest LPIPS metrics among all compared methods, while reaching second-best position on the PSNR metric (see Table \ref{lol1}). Furthermore, Table \ref{lol2} reports benchmark metrics on \texttt{LOL-v2-Real} and \texttt{LOL-v2-Synthetic}, clearly showing once again that our method achieves state-of-the-art performance in terms of SSIM and LPIPS, while systematically reaching second-best performance in terms of PSNR. 

Overall, when evaluating the average SSIM, PSNR and LPIPS across all datasets, Table \ref{averages} clearly shows that our method outperforms prior approaches by reaching the highest average SSIM, PSNR and LPIPS measures.

\begin{table}[htbp]
\centering
\small
\setlength{\tabcolsep}{6pt}
\caption{Average benchmark metrics across \texttt{LOL-v1}, \texttt{LOL-v2-Real} and \texttt{LOL-v2-Synthetic}.}
\begin{tabular}{lccc}
\hline
Method & SSIM$\uparrow$ & PSNR$\uparrow$ & LPIPS$\downarrow$ \\
\hline

LIME~\cite{Guo2016LIME} & 0.603 & 16.26 & 0.480 \\
SRIE~\cite{Fu2016WeightedVariational} & 5.431 & 10.40 & 0.289 \\
BVIF~\cite{Yang2020BVIF} & 0.608 & 16.76 & 0.267 \\
BIMEF~\cite{Ying2017BIMEF} & 0.598 & 15.13 & 0.275 \\
ZeroDCE++~\cite{Li2021ZeroDCEpp} & 0.665 & 17.84 & 0.272 \\
EnlightenGAN~\cite{Jiang2021EnlightenGAN} & 0.667 & 17.43 & 0.281 \\
RetinexNet~\cite{Wei2018RetinexNet} & 0.642 & 16.46 & 0.531 \\
KinD~\cite{Zhang2019KinD} & 0.670 & 16.30 & 0.339 \\
MIRNet~\cite{Zamir2020MIRNet} & 0.842 & 22.03 & 0.187 \\
UFormer~\cite{Wang2022UFormer} & 0.804 & 18.28 & 0.258 \\
Restormer~\cite{Zamir2022Restormer} & 0.827 & 21.26 & 0.133 \\
LANet~\cite{Yang2023LANet} & 0.824 & 21.15 & 0.116 \\
SurroundNet~\cite{Zhou2023SurroundNet} & 0.853 & 22.62 & 0.098 \\
Retinexformer~\cite{Cai2023Retinexformer} & 0.853 & 23.09 & 0.128 \\

SCFR-WLMB~\cite{lv2025low} & 0.854 & 23.51 & -- \\
LightenDiffusion~\cite{jiang2024lightendiffusion} & 0.844 & 21.44 & 0.167 \\

\hline
\textbf{IAML (ours)} & \textbf{0.888} & \textbf{23.37} & \textbf{0.098} \\
\hline

\end{tabular}
\label{averages}
\end{table}

\begin{table}[htbp]
\centering
\small
\setlength{\tabcolsep}{6pt}
\caption{Ablation study on the effect of the loss formulation on the average SSIM, PSNR and LPIPS. Best results are shown in bold, and second-best results are \underline{underlined}.}
\begin{tabular}{lccc}
\hline
Loss Configuration & SSIM$\uparrow$ & PSNR$\uparrow$ & LPIPS$\downarrow$ \\
\hline
1) MSE only 
& 0.8545 & 23.3342  & 0.0935 \\

2) MSE + SSIM 
& 0.8722 & 23.1278  & 0.0897 \\



3) MSE + SSIM + Cos. Sim. 
& 0.8606 & 23.3549  & 0.0989 \\

4) MSE + SSIM + Std. $\ell_1$ 
& \underline{0.8737} & \underline{23.801}  & \underline{0.0882} \\

5) \textbf{MSE + SSIM + IAML} 
& \textbf{0.888} & \textbf{23.37}  & \textbf{0.098}  \\

\hline
\end{tabular}
\label{ablation}
\end{table}

\subsection{Ablation studies}

Importantly, in order to explore the impact of our proposed IAML approach, we perform ablation studies on the mirror loss formulation in (\ref{eq:mirror}) and on the total loss formulated in (\ref{eq:total}). Table \ref{ablation} reports the SSIM, PSNR and LPIPS metrics for five different ablation cases on our loss formulation. In Table \ref{ablation}, entry 3) replaces the IAML with \textit{cosine similarity} losses between student and teacher decoder feature maps. Furthermore, entry 4) replaces the IAML with an $\ell_1$ loss between standardized student and teacher decoder feature maps (\ref{eq:instnorm}). Finally, entry 5) denotes our full IAML approach. Table \ref{ablation} clearly demonstrates that our custom IAML formulation (entry 5) leads to thee best performance, showing the crucial effect of illumination-aware feature map distillation during training.

\section{Conclusion}
\label{concss}
This paper has presented a novel method for the training of low-light image enhancement auto-encoders. The proposed IAML framework relies on a teacher-student pair of auto-encoder network where information from teacher feature maps corresponding to clean input image representations are used to guide the training of the student model, by aligning the student decoder's internal representations with the teacher ones. Extensive benchmarking and ablation studies have clearly demonstrated the usefulness of our approach, outperforming more than a dozen prior methods in terms of average SSIM, PSNR and LPIPS across three datasets. As future work, we plan to apply our approach to other image restoration tasks in order to study its validity across various application domains.

\ifCLASSOPTIONcaptionsoff
  \newpage
\fi



%
\bibliographystyle{IEEEtran}
\bibliography{cas-refs}

%








\end{document}